\documentclass[12pt,twocolumn,american]{article}
\usepackage{amsmath}
\usepackage{amsthm}
\usepackage{fontspec}
\usepackage{geometry}
\usepackage{float}
\usepackage{graphicx}
\usepackage{esint}
\usepackage{microtype}
\usepackage[unicode=true,
 bookmarks=true,bookmarksnumbered=false,bookmarksopen=false,
 breaklinks=false,pdfborder={0 0 1},backref=false,colorlinks=false]
 {hyperref}
\hypersetup{pdftitle={Monitoring Gallium Flow with Event Camera: Parameter Optimization and Measures Characterization},
 pdfauthor={R. Tamir, D. Ben-Ezra, D. Brisk, R. Arad, S. Vaintraub}}

\makeatletter

\providecommand\textquotedblplain{%
  \bgroup\addfontfeatures{RawFeature=-tlig}\char34\egroup}

\numberwithin{equation}{section}
\numberwithin{figure}{section}

\@ifundefined{date}{}{\date{}}
\makeatother

\usepackage{polyglossia}
\setdefaultlanguage[variant=american]{english}
\begin{document}
 \twocolumn[
   \begin{@twocolumnfalse}
\title{Monitoring a Liquid Gallium--Indium Jet Flow Using an Event Camera}
\author{Raz-li Tamir, David El-Chai Ben-Ezra, Sergei \\
Vaintraub, Ilan Eliyahu, Daniel Brisk, Ilya Lapin}
\maketitle
\begin{abstract}
This paper establishes the foundation for optical monitoring of gallium--indium
flow at the Soreq Applied Research Accelerator Facility (SARAF-II)
beam dump using an event camera. We introduce a novel paradigm for
event-based optical flow monitoring that leverages the asynchronous
sensing capabilities of neuromorphic vision technology. The study
defines global and regional performance metrics based on the event
rate generated by the camera. Within this framework, the physical
gallium--indium flow is transformed into an “event flow,” whose dynamics
reflect the underlying flow behavior. Because event cameras are relatively
new and highly distinctive sensing devices, tuning and optimizing
their bias parameters for a specific application remains a challenging
task. Accordingly, we also present a heuristic approach for evaluating
and comparing different bias configurations in order to obtain an
optimal signal for flow monitoring.
\end{abstract}

 \end{@twocolumnfalse}  ]

\section{Introduction\label{sec:Introduction}}

\begin{figure}[H]
\centering\includegraphics[scale=0.6]{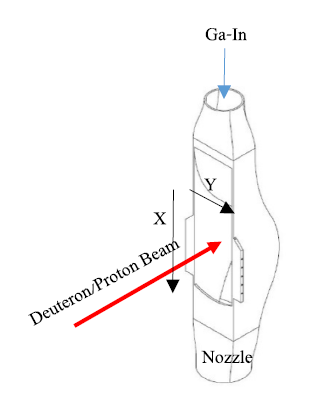}\caption{The GaIn nozzle.}

\label{fig 1.1}
\end{figure}

Soreq Applied Research Accelerator Facility (SARAF-II) is based on
a proton/deuteron linear accelerator \cite{key-1}. To enable efficient
heat transfer, a liquid gallium--indium (GaIn) metal alloy target,
with a melting point of approximately $15^{\circ}\,C$, was designed
to absorb a beam with a maximum power of $200\,kW$. The system requires
a stable liquid film approximately $6\,mm$ thick, flowing at a velocity
of about $5\,m/s$ along a concave supporting wall through a nozzle
with dimensions of roughly $5\times10\,cm^{2}$ (see Figure \ref{fig 1.1})
\cite{key-2}.

To prevent GaIn boiling and excessive surface evaporation rate, the
GaIn flow velocity must remain stable with minimal lateral undulations
or other irregularities. One method for monitoring the flow is to
employ a high-speed optical sensor positioned to observe the GaIn
stream directly through the nozzle and detect flow anomalies in real
time. A similar approach was implemented in the EVEDA accelerator
\cite{key-3}, where a $20\,kHz$ high-speed camera was used \emph{for}
flow monitoring.

An event camera is a bio-inspired vision sensor that captures changes
in a scene rather than recording full image frames. Each pixel operates
asynchronously and independently, generating a response whenever the
local brightness variation exceeds a predefined threshold. The camera
output consists of a stream of timestamped events, where each event
contains the pixel coordinates, a timestamp with microsecond resolution,
and the polarity of the brightness change. In the absence of motion
or intensity variations within the field of view, the output of an
event camera is ideally empty. This differs fundamentally from conventional
frame-based cameras, which continuously produce full image frames
regardless of scene activity. By transmitting only relevant changes,
event cameras naturally support the detection of fast dynamics and
high-speed motion without requiring high frame rates over the entire
image. Consequently, they are particularly well suited for applications
that traditionally rely on high-speed imaging, such as monitoring
the GaIn flow.

Optical monitoring of flow requires appropriate reference metrics
and image-processing techniques for detecting anomalies. A traditional
approach to optical flow measurement is Particle Image Velocimetry
(PIV), which typically involves introducing small tracer particles
into the fluid and tracking their motion using cross-correlation methods.
For example, \cite{key-4} presents a PIV approach using an event
camera, where particle-induced events are aligned, and fitting a plane
in the spatiotemporal $x-y-t$ domain yields optical flow information
without the need for conventional frame-based imaging (see also \cite{key-5}-\cite{key-10}
for this approach). Even so, PIV can also be applied to liquid metal
flows without seeding particles, as demonstrated in \cite{key-3},
where high-speed imaging of natural light patterns in liquid lithium
enabled flow tracking. 

However, the new paradigm introduced by event cameras motivates the
development of novel approaches for flow monitoring. The method presented
in this paper leverages the unique output of an event camera to represent
the GaIn flow as a \textbf{stream of events}. Unlike conventional
flow-monitoring techniques, which rely primarily on image processing,
our principal benchmark is the \textbf{event-rate} (see $\oint$\ref{sec:Primary-Measures-and}).
Variations in the event rate directly reflect changes in the GaIn
flow, providing an immediate and efficient response to flow dynamics.
For example, since event cameras respond only to changes in illumination
at each pixel, an abrupt halt in the GaIn flow results in a corresponding
cessation of event generation. Additionally, as demonstrated in $\oint$\ref{sec:Demonstration},
the event rate and its derivative benchmarks closely follow variations
in the GaIn flow, enabling a deeper understanding of anomalous flow
behavior.

Event cameras require tuning several bias parameters (see $\oint$\ref{sec:Event-Camera:-Background}),
which differ significantly from conventional frame-based camera settings
such as frame rate and exposure time. Determining the optimal configuration
of these biases to fully leverage the camera’s capabilities for a
given task is a nontrivial challenge. Accordingly, a central focus
of this work is the development of appropriate signal benchmarks and
an accompanying optimization methodology (see $\oint$\ref{sec:Tuning-Biases-and}).

\section{Preliminaries \label{sec:Event-Camera:-Background}}

In contrast to conventional cameras, an event camera does not operate
with a fixed frame rate. Rather than uniformly sampling the entire
field of view at regular intervals, each pixel functions asynchronously
and independently, responding only to changes in its own brightness
level. When such a change is detected, the pixel generates an event
of the form 

\[
e_{k}=\left(t_{k},x_{k},y_{k},p_{k}\right)
\]
where $t_{k}$ represents the timestamp (in microseconds), $x_{k},y_{k}$
denote the pixel coordinates, and $p_{k}$ indicates the polarity---taking
a value of $+1$ for an increase in brightness and $-1$ for a decrease.
The output of an event camera is thus a temporally ordered stream
of events $e_{k}$. For further background, see \cite{key-11}.

\begin{figure}[H]
\centering\includegraphics[scale=0.8]{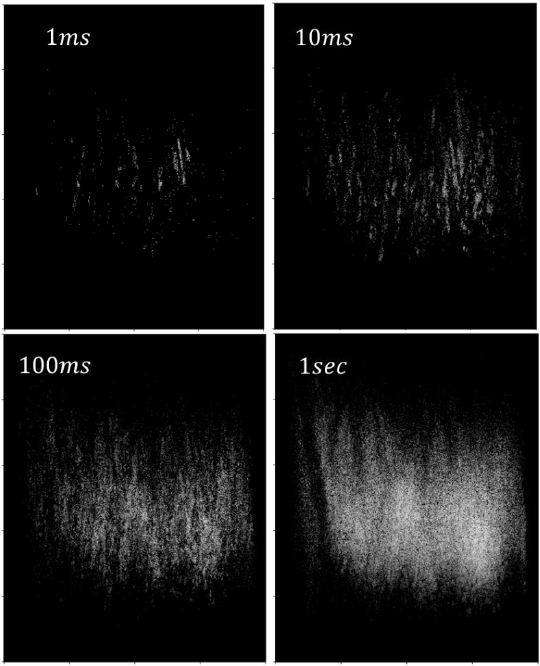}\caption{Synthetic event frames of the GaIn flow over time intervals ranging
from $1\,ms$ to $1\:s$.}

\label{fig 1.1-1}
\end{figure}

Figure \ref{fig 1.1-1} illustrates synthesized “frames” constructed
from events recorded by the camera during GaIn flow measurements,
accumulated over varying time intervals. In this representation, pixel
brightness corresponds to the relative number of events generated
by each pixel. It can be observed that, due to the lighting conditions
and the specifics of the measurement setup, the region producing events
is confined to less than half of the approximately $\sim5\times10\,cm^{2}$
observation window.

A key challenge in working with event cameras is tuning the parameters
that govern event generation. Each device includes several biases
that must be carefully adjusted to suit the specific task in order
to achieve optimal performance. In this study, we employed the EVK4-HD
event camera by Prophesee, currently one of their most advanced models.
This camera provides several configurable parameters:
\begin{itemize}
\item \textbf{bias\_diff\_on:} sets the threshold for an increase in illumination
required to trigger a positive event.
\item \textbf{bias\_diff\_off:} sets the threshold for a decrease in illumination
required to trigger a negative event.
\item \textbf{bias\_fo:} defines a low-pass filter at the pixel level, limiting
sensitivity to rapid light fluctuations.
\item \textbf{bias\_hpf:} defines a high-pass filter at the pixel level,
suppressing slow variations in illumination. 
\item \textbf{bias\_refr:} determines the refractory period between consecutive
events generated by the same pixel. The default value in the EVK4-HD
ranges from $10\,\mu s$ to $100\,\mu s$ \cite{key-12}, which is
already sufficiently low; therefore, it was not modified in this work.
\end{itemize}
These parameters do not have fixed absolute meanings, but they exhibit
several consistent properties: all parameters have a default value
of $0$; decreasing the values of bias\_diff\_on and bias\_diff\_off
increases the camera’s sensitivity to brightness changes; and increasing
bias\_hpf and bias\_fo shifts the camera’s response toward higher
frequencies---specifically, a higher bias\_hpf suppresses more low-frequency
components, while a higher bias\_fo allows more high-frequency variations
to pass.

\section{Measurement Methodology\label{sec:Tuning-Biases-and}}

\begin{figure*}[t]
\centering\includegraphics[scale=0.5]{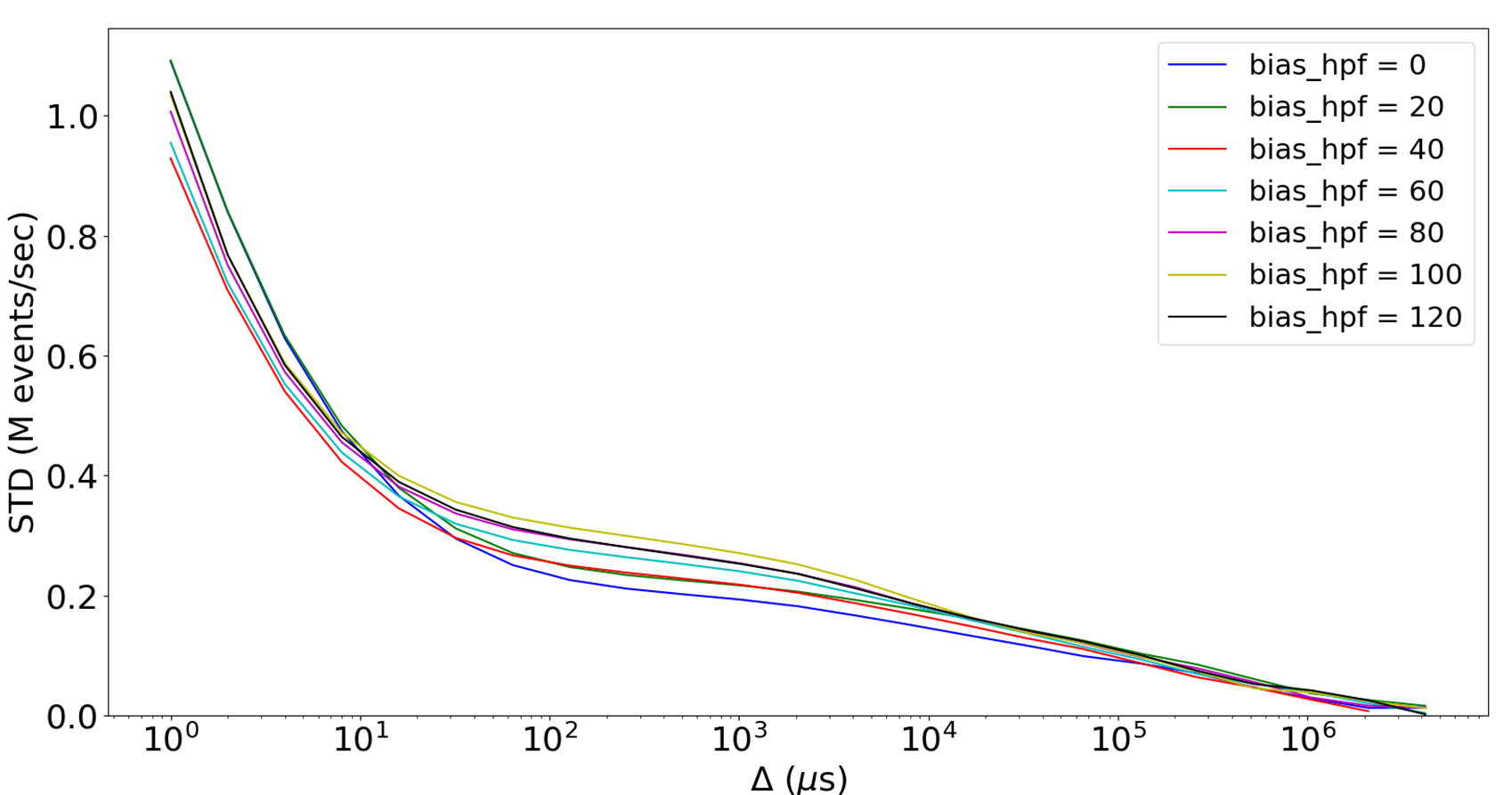}

\caption{Standard deviation of the event rate as a function of the division
constant $\Delta$, with each curve corresponding to a different bias\_hpf
value used during recording.}

\label{fig 3.2}
\end{figure*}

\subsection{Sensitivity Tuning}

In this work, we adopt the sensitivity-tuning heuristic introduced
in \cite{key-13}, preserving two key principles: maintaining a fixed
global event rate and achieving a balanced ratio between positive
and negative events. Contrary to intuition, event cameras are not
inherently calibrated to produce equal numbers of positive and negative
events, and achieving such balance requires careful tuning. To ensure
both a stable event-rate and a balanced polarity distribution, we
employed the algorithm used in \cite{key-13} to adjust the sensitivity
parameters. Specifically, the algorithm optimizes bias\_diff\_on and
bias\_diff\_off to approximate equality between positive and negative
event counts while maintaining a target event rate per second. This
procedure was applied at the beginning of each experiment, setting
the global event rate to approximately $1\,M$ $events/sec$ under
standard flow conditions, after which variations in event rate and
other benchmarks were analyzed during recording. 

\begin{figure*}[t]
\centering\includegraphics[scale=0.5]{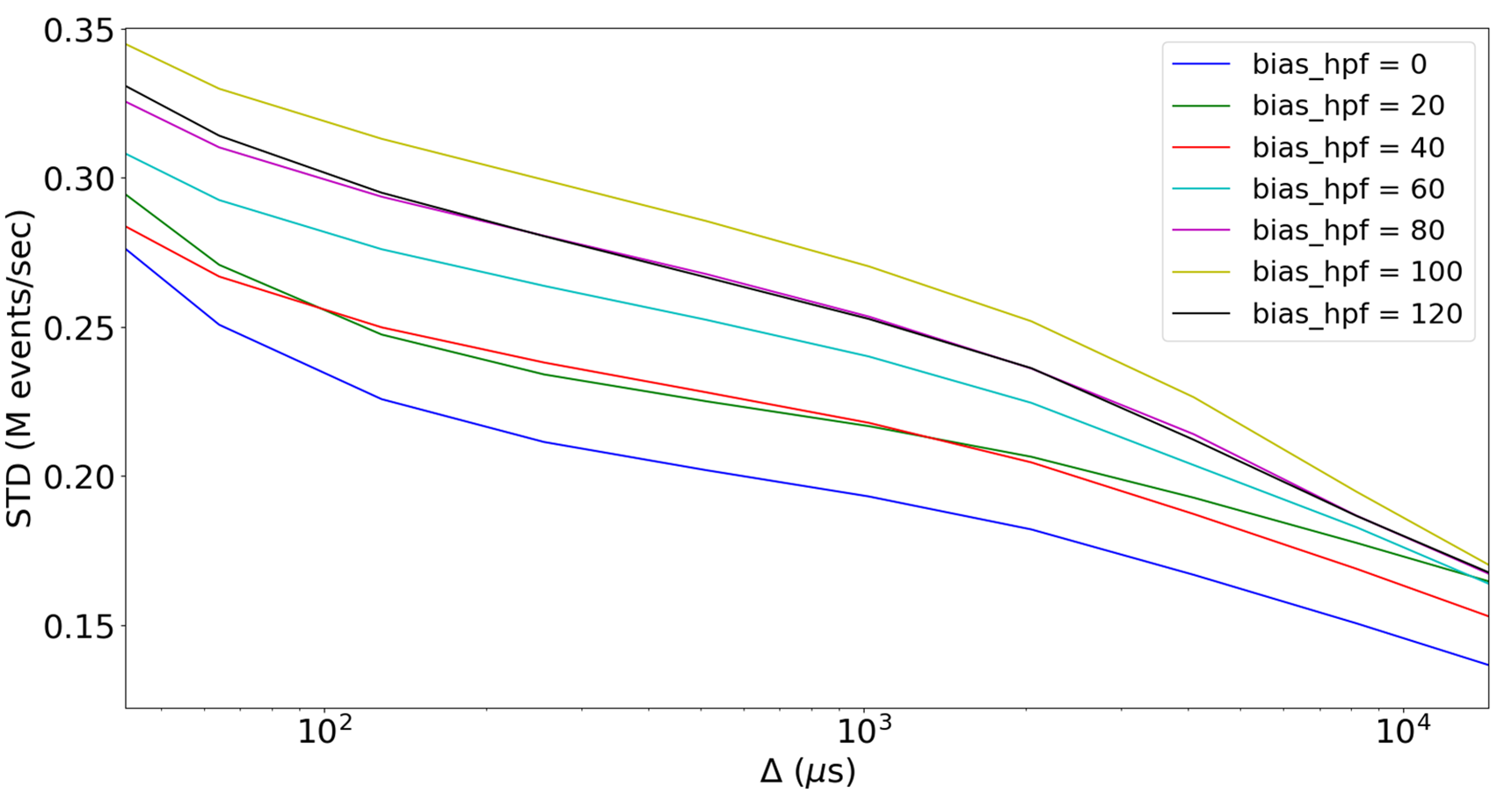}\caption{Standard deviation as a function of the division constant $\Delta$,
shown as a close-up of Figure \ref{fig 3.2} over the range $10^{2}\,\mu s\protect\leq\Delta\protect\leq10^{4}\,\mu s$.}

\label{fig 3.3}
\end{figure*}

\subsection{The Division Constant}

As noted in the introduction, our primary benchmark is the event-rate.
To compute it, we partition the time axis into intervals of equal
duration, denoted by $\Delta$, referred to as the time division constant.
Within each interval, we count the number of events that occur, forming
a sequence that represents the event-rate for the chosen $\Delta$.
Selecting an appropriate value for $\Delta$ is crucial for obtaining
a meaningful signal: if the interval is too large, important features
may be smoothed out, causing the monitoring system to respond slowly
to variations in the GaIn flow; conversely, if $\Delta$ is too small,
the resulting signal becomes noisy and may lead to unstable monitoring.

To determine an appropriate and balanced choice for $\Delta$, we
evaluated the standard deviation (STD) of the event rate as a function
of $\Delta$ across seven recordings, each obtained with different
bias\_hpf values ranging from $0$ to $120$ (the default minimum
and maximum settings). In all experiments, bias\_fo was kept at its
default value of $0$. We focused on bias\_hpf rather than bias\_fo,
as prior work \cite{key-13} suggests that it has a more pronounced
effect on the signal behavior. While we expected the STD of the event
rate to decrease with increasing $\Delta$, we also anticipated a
transition point at which $\Delta$ begins to act as a low-pass filter,
suppressing important signal features. The intuition behind this approach
is illustrated in $\oint$\ref{subsec:Results:-Comparing-Biases}.
The results of this analysis are presented in Figure \ref{fig 3.2}.

It can be observed that all curves in Figure \ref{fig 3.2} exhibit
an inflection point in the range $1\,ms\leq\Delta\leq10\,ms$. In
view of this observation, we set $\Delta=10\,ms$ for all subsequent
demonstrations, for two main reasons:
\begin{itemize}
\item \textbf{Physical consideration:} At $\Delta=10\,ms$, the signal features
of interest remain preserved, ensuring sensitivity to relevant flow
dynamics. Notably, this timescale closely matches the time required
for GaIn flowing at approximately $5\,m/s$ to traverse the illuminated
region of about $5\,cm$ within the $10\,cm$ observation window (see
Figure \ref{fig 4.1}).
\item \textbf{Technical consideration:} Event cameras periodically transmit
accumulated events in packets. In the Prophesee EVK4-HD model, the
default packet interval is $40\,ms$, though this can be adjusted.
Increasing this interval slows response time, while decreasing it
increases packet frequency, with uncertain implications for real-time
performance. Setting the packet interval equal to $\Delta$ simplifies
event-rate computation, as it allows direct counting of events per
packet. We estimate that reducing the interval to $10\,ms$ provides
a good balance between computational simplicity and system responsiveness,
while avoiding potential degradation associated with higher transmission
rates.
\end{itemize}

\subsection{High-Pass Filter Optimization}

Figure \ref{fig 3.3} presents a magnified view of Figure \ref{fig 3.2}
over the interval $10^{2}\,\mu s\leq\Delta\leq10^{4}\,\mu s$. A clear
ordering can be observed among the curves corresponding to different
bias\_hpf values: lower bias\_hpf settings produce lower standard
deviation values, while increasing bias\_hpf generally leads to higher
standard deviations. An optimum is observed at bias\_hpf $=100$,
beyond which the standard deviation decreases, as seen for bias\_hpf
$=120$. As discussed in $\oint$\ref{subsec:Results:-Comparing-Biases},
a relatively high standard deviation indicates greater sensitivity
to flow features. Based on this observation, we determined that the
optimal bias\_hpf value for monitoring GaIn flow is $100$, and this
setting is therefore used in the demonstration presented in $\oint$\ref{sec:Demonstration}.

\subsection{Intuition Behind the STD Benchmark \label{subsec:Results:-Comparing-Biases}}

\begin{figure*}[t]
\centering\includegraphics[scale=0.5]{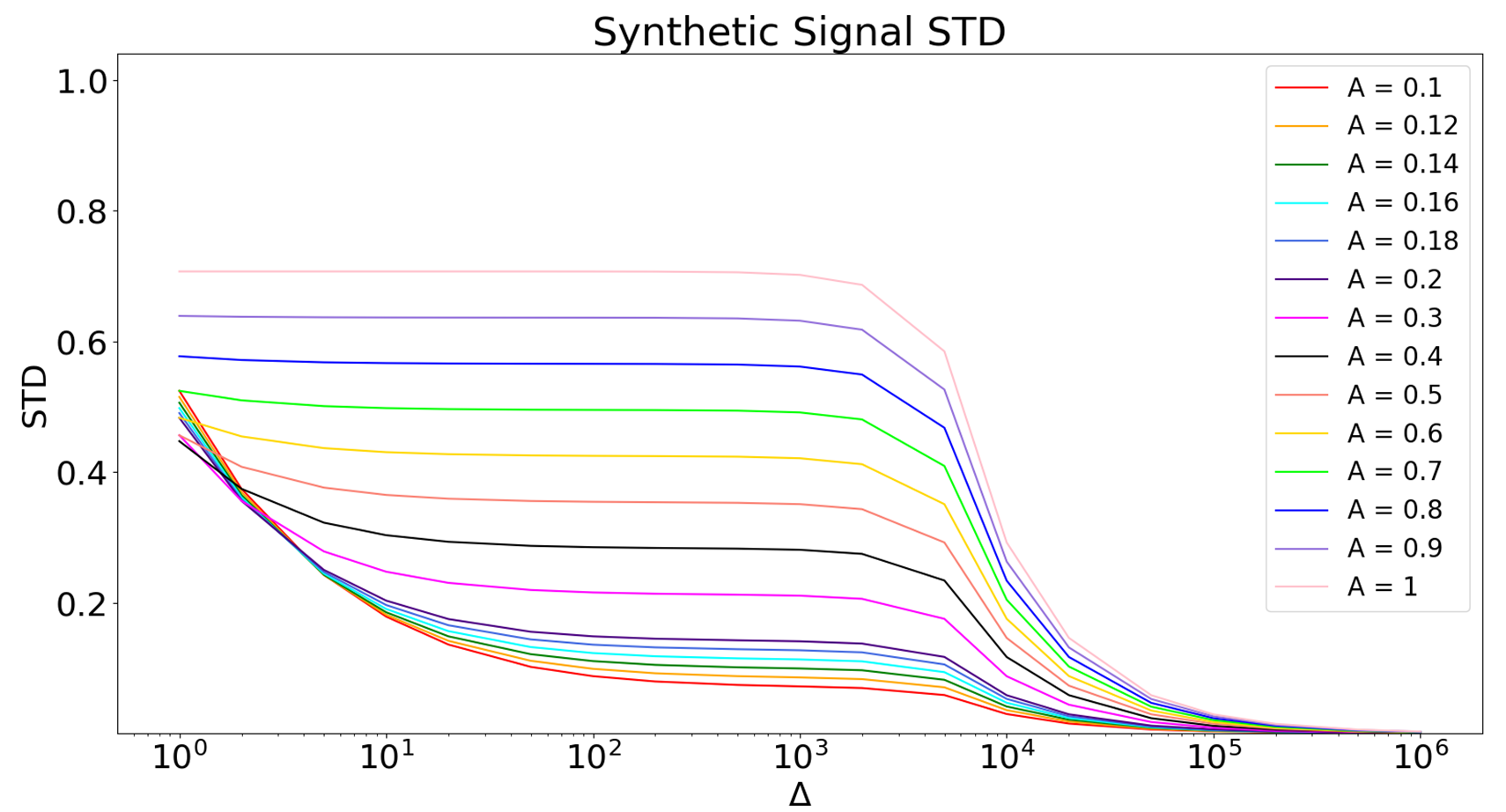}

\caption{Standard deviation of a synthetic periodic signal combined with random
noise. Each plot represents a different weight $A$ of the periodic
component.}

\label{fig 3.1}
\end{figure*}

Figure \ref{fig 3.1} illustrates the interpretation of the inflection
points shown in Figure \ref{fig 3.2}, which motivate the selection
of $\Delta$, as well as the interpretation of the hierarchy observed
among the graphs in Figure \ref{fig 3.3}. The figure displays the
standard deviation (STD) of synthetic signals constructed as follows.

Consider a sinusoidal function with period $T$, shifted upward so
that it remains non-negative: 
\[
\sin\left(\dfrac{2\pi t}{T}\right)+1.
\]
This function has a non-negative range, and its average over a sufficiently
long interval is approximately $1$. By adding a weighted uniformly
distributed random signal $R\sim U\left(0,2\right)$, we obtain 

\[
f\left(t\right)=A\cdot\left(\sin\left(\dfrac{2\pi t}{T}\right)+1\right)+\left(1-A\right)\cdot R
\]
where $0\leq A\leq1$ determines the contribution of the sinusoidal
component. The resulting function $f(t)$ remains non-negative and
maintains an average value of approximately $1$. We use $f(t)$ as
an analogy to the event-rate signal observed during GaIn flow: both
are non-negative and exhibit a roughly constant average. The sinusoidal
term represents structured, deterministic features in the signal,
while the random component reflects stochastic variations in the flow.
We emphasize that this construction is not intended to precisely model
the event-rate behavior of GaIn flow, but rather to illustrate how
the standard deviation behaves for signals with a similar conceptual
structure.

Figure \ref{fig 3.1} shows the standard deviation (STD) of $f$ for
$T=15\cdot10^{-3}$, with $A$ varying in the range $0.1\leq A\leq1$,
and includes a zoomed-in view for $0.1\leq A\leq0.2$. The axes are
presented in arbitrary units to emphasize the synthetic nature of
the signal, although the horizontal axis can be interpreted as corresponding
to microseconds, analogous to Figure \ref{fig 3.2}. The choice $T=15\cdot10^{-3}$
was made to align the inflection point with that observed in Figure
\ref{fig 3.2}, indicating that the non-random features of the GaIn
event-rate signal become evident when using $\Delta=10\,ms$. Furthermore,
the graphs in Figure \ref{fig 3.2} resemble those in Figure \ref{fig 3.1}
for the range $0.1\leq A\leq0.2$, suggesting that the GaIn event-rate
signal is predominantly random. Finally, the ordering of the curves
in Figure \ref{fig 3.1} indicates that a higher STD corresponds to
a stronger non-random component, supporting the conclusion that a
bias setting of bias\_hpf $=100$ is optimal for preserving the weak
structured features of the event-rate signal.

\section{Monitoring Benchmarks\label{sec:Primary-Measures-and}}

In this section, we define basic benchmarks for monitoring the GaIn
flow. These benchmarks can be used either as standalone monitoring
criteria or as a foundation for developing more advanced measures.
As noted in the previous section, the primary signal is the event
rate, obtained by counting events after dividing the time axis into
intervals of duration $\Delta=10\,ms$. 

\subsection{Global Event Rate}

As noted in $\oint$\ref{sec:Tuning-Biases-and}, evaluating the event
rate over intervals of $\Delta=10\,ms$ preserves the non-random characteristics
of the signal. However, to also monitor the average event flow, it
is useful to consider longer time intervals, as they reduce sensitivity
to fluctuations. We achieve this by computing the exponential moving
average (EMA) of the original signal: 

\[
MA\left[t\right]=\alpha\cdot rate\left[t\right]+\left(1-\alpha\right)\cdot MA\left[t-1\right]
\]
where:
\begin{itemize}
\item $MA\left[t\right]$ is the exponential moving average at time $t$,
\item $rate[t]$ is the event rate at time $t$, 
\item $0<\alpha\leq1$ is the weighting factor assigned to each new sample.
\end{itemize}
Setting $\alpha=1$ reduces the expression to the original signal.
All previous samples are recursively weighted by $\left(1-\alpha\right)$.
We examined several values of $\alpha=0.1,0.01,0.001$. Notably, using
$\alpha=10^{-n}$ is roughly equivalent to averaging over a time window
of $\sim\Delta\cdot10^{n}$. Thus, $\alpha=0.1$ corresponds to an
averaging window of approximately $100\,ms$, $\alpha=0.01$ to about
$1\,s$, and $\alpha=0.001$ to roughly $\sim10\,s$.

Note that the exponential moving average can be computed naturally
in real time, as it only requires the current event-rate measurement
for the latest event packet and the previous EMA value. In particular,
there is no need to store past packets to perform the calculation.
Moreover, it offers an advantage over the standard moving average
in that the weighting factor can be adjusted continuously. This allows
effective evaluation of the event rate over arbitrary time scales,
without being restricted to multiples of $\Delta$.

\subsection{Local Monitoring \label{subsec:Local-Monitoring}}

\begin{figure}[H]
\centering\includegraphics{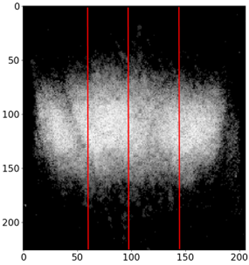}

\caption{Event map partitioned into four strips according to the active pixel
map, computed over a one-second interval.}

\label{fig 4.1}
\end{figure}

In addition to the global monitoring measures, we also perform local
monitoring in order to detect subtle variations that may not be visible
at the global level. This is done by dividing the signal area into
four strips and evaluating the event rate for each strip separately,
using the same approach as in the global analysis. The partition into
strips is illustrated in Figure \ref{fig 4.1}.

As shown in the figure, lighting effects cause the edges of the flow
to exhibit significantly lower activity compared to the center. Therefore,
dividing the area into equally spaced strips and comparing their event
rates would result in imbalanced values across the strips. To address
this, we adopt a different partitioning approach: first, we compute
the active pixel map over a one-second interval. We then test various
partitions and select the one that produces the most balanced distribution
of active pixels across the regions. This procedure is subsequently
applied again to each half. The resulting partition yields areas with
approximately equal event rates.

\section{Flow Monitoring\label{sec:Demonstration}}

\begin{figure*}[tp]
\centering\includegraphics[scale=0.5]{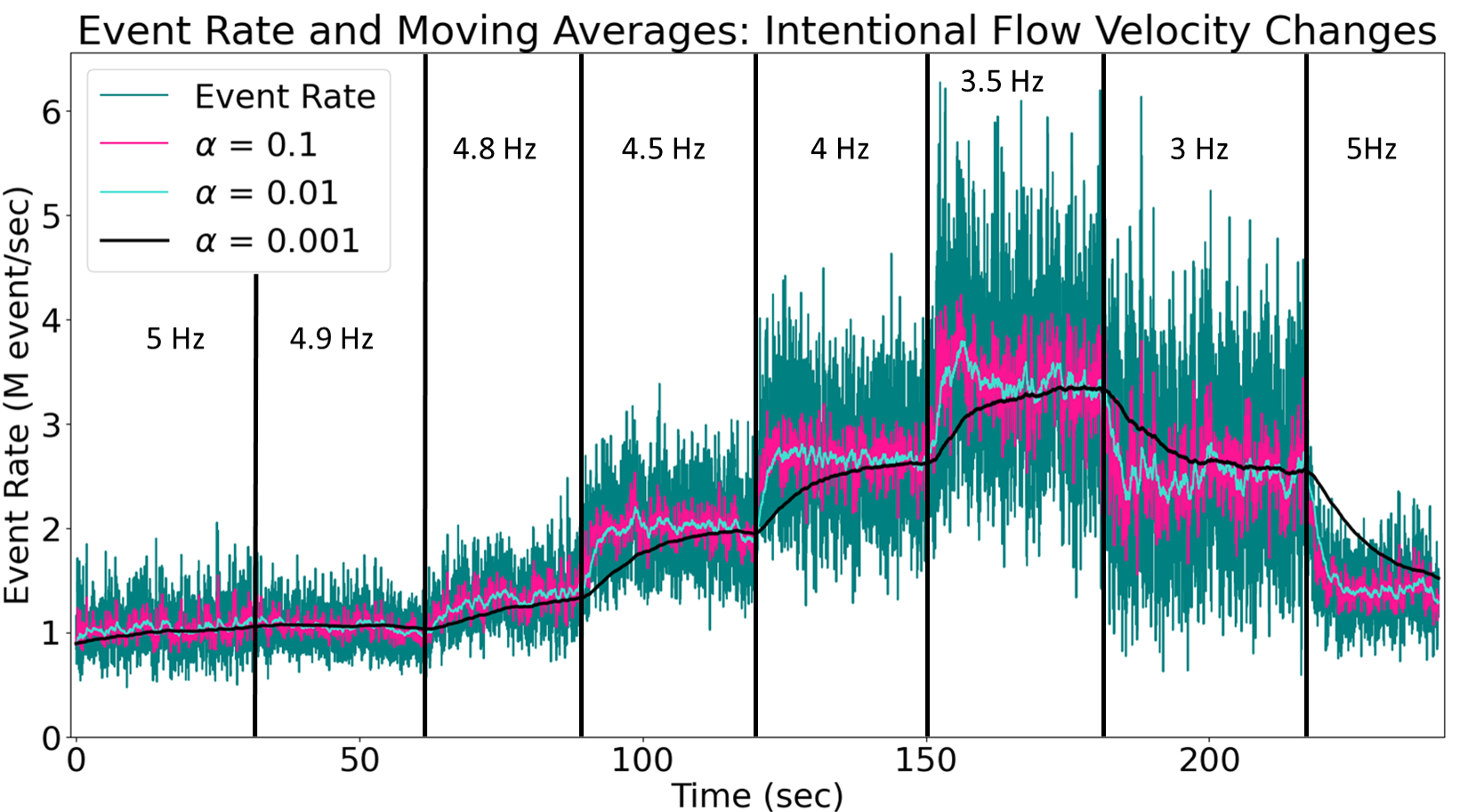}\caption{Event rates and exponential moving averages for the intentional velocity-change
experiment. The moving averages were computed using smoothing factors
$\alpha=0.1,0.01,0.001$.}

\label{fig 5.1}
\end{figure*}

\begin{figure*}[p]
\centering\includegraphics[scale=0.5]{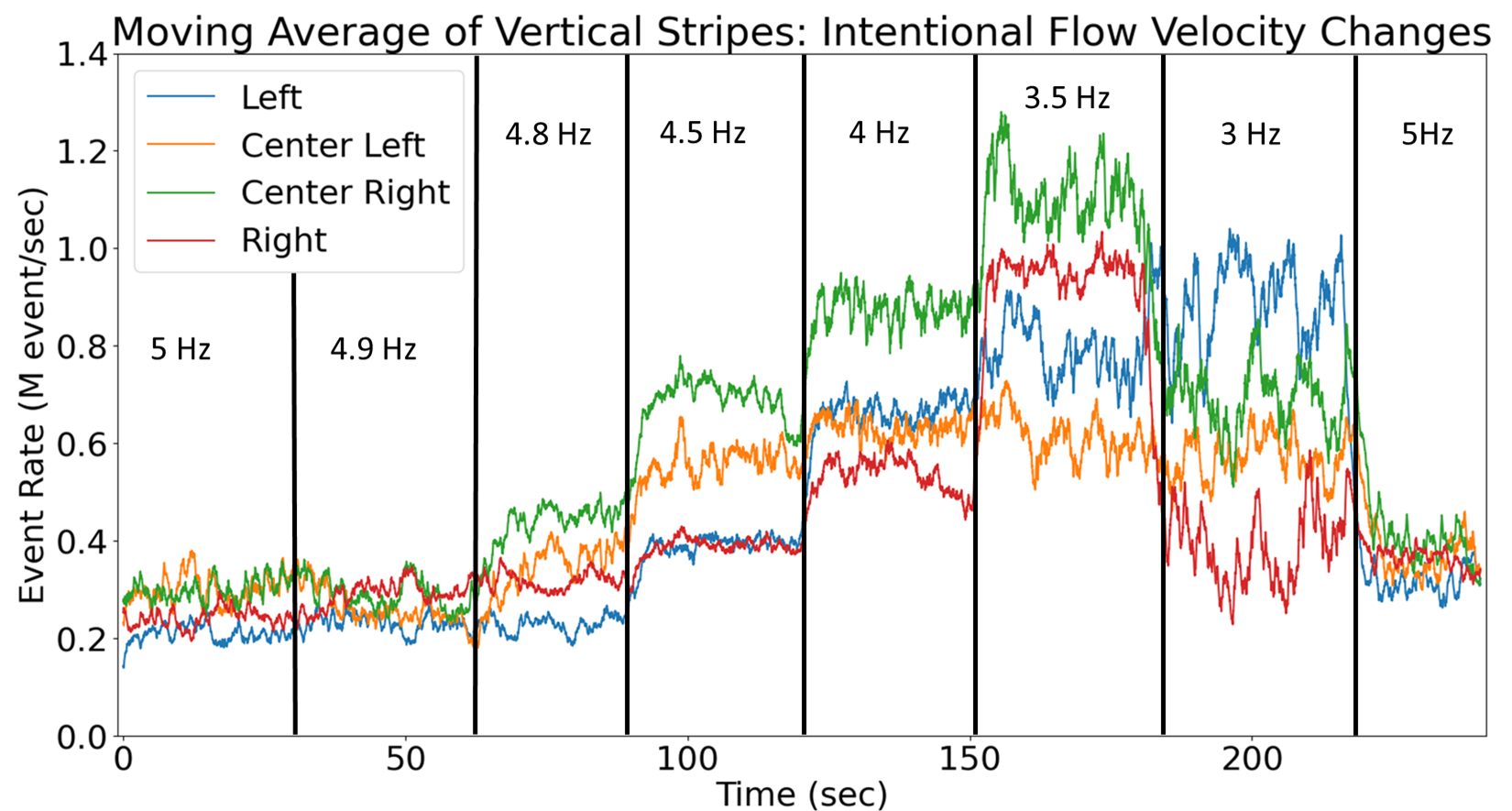}\caption{Exponential moving averages ($\alpha=0.01$) for the four vertical
stripes under the intentional flow velocity‑change scenario.}

\label{fig 5.2}
\end{figure*}

In this section, we demonstrate the monitoring heuristic developed
in two scenarios. In the first scenario, the flow velocity was intentionally
modified by reducing the speed approximately every 30 seconds, and
the response of the benchmarks described in $\oint$\ref{sec:Primary-Measures-and}
was analyzed (see $\oint$\ref{subsec:Scenario-1}). In the second
scenario, a stable flow was maintained for about three minutes, during
which the behavior of the same benchmarks was monitored over time.
As will be shown, the results reveal subtle variations that indicate
potential anomalies (see $\oint$\ref{subsec:Scenario-2:-Constant}). 

In both experiments, the nominal velocity for stable GaIn flow was
set to $5\,Hz$. Here, velocity is expressed in $Hz$ rather than
$m/s$ due to constraints related to the GaIn electromagnetic pump.
Following the procedure described in $\oint$\ref{sec:Tuning-Biases-and},
the data were recorded with a target event rate of $1\,M\,events/s$,
$\Delta=10\,ms$, bias\_hpf $=100$ , and bias\_fo $=0$.

\begin{figure*}[t]
\centering\includegraphics[scale=0.5]{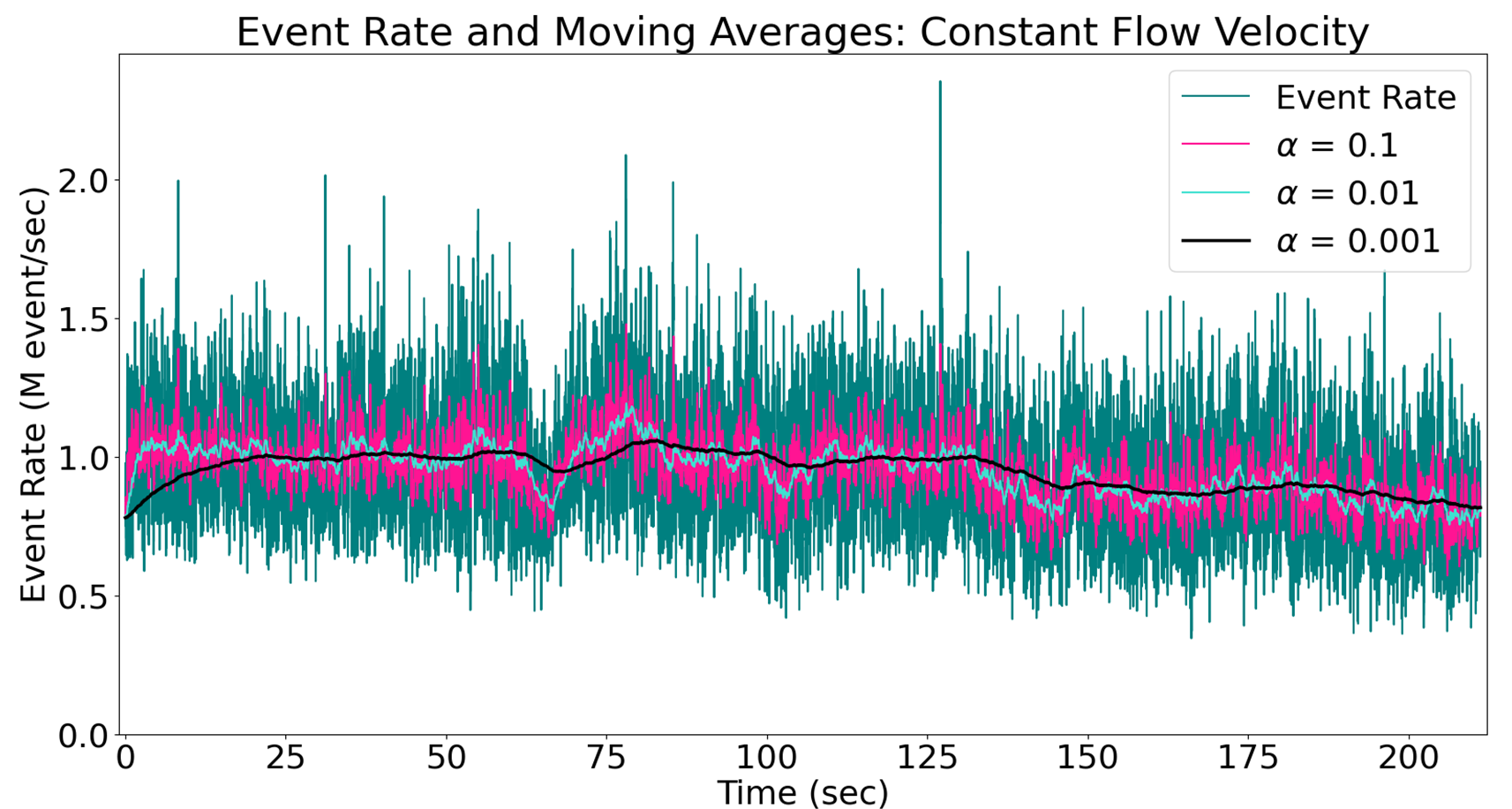}

\caption{Event rates and moving averages for a recording with constant flow
velocity. A slight decline in the global event rate is observable
after approximately 130 seconds.}

\label{fig 5.3}
\end{figure*}

\subsection{Scenario 1: Deliberate Variations in Flow Velocity\label{subsec:Scenario-1}}

Starting from the standard baseline flow velocity of $5\,Hz$, the
velocity was adjusted approximately every 30 seconds. The sequence
of changes was:

\begin{align*}
5\,Hz & \rightarrow4.9\,Hz\\
 & \rightarrow4.8\,Hz\\
 & \rightarrow4.5\,Hz\\
 & \rightarrow4\,Hz\\
 & \rightarrow3.5\,Hz\\
 & \rightarrow3\,Hz\\
 & \rightarrow5\,Hz.
\end{align*}

Figure \ref{fig 5.1} illustrates how these variations in GaIn flow
affect the event stream. Under steady conditions at the default velocity
of $5\,Hz$, the event rate is approximately 1 million events per
second ($1\,M\,events/s$). Although it may initially seem counterintuitive,
reducing the flow velocity, up to about $3.5\,Hz$, leads to an increase
in the event rate. At $3\,Hz$, however, the event rate decreases
slightly while exhibiting strong oscillations. This behavior can be
explained by the flow dynamics: at $5\,Hz$, the flow is most stable
and continuous. Lowering the velocity introduces turbulence and disrupts
continuity, resulting in the formation of distinct droplets. These
discontinuities increase both the event rate and its variability.
After returning to the default velocity of $5\,Hz$, the event rate
remains somewhat elevated compared to its initial level. This effect
likely arises from the abrupt velocity changes, as the system and
flow require additional time to fully return to a stable state.

This scenario clearly illustrates the concept of “event‑flow” introduced
in $\oint$\ref{sec:Introduction}. Variations in the GaIn flow lead
to pronounced changes in the event rate: deviations from the stable
flow velocity immediately produce an increase in event activity, and
the event rates corresponding to different velocities remain distinct
from one another. Notably, this event‑flow characterization is computationally
efficient, as it does not rely on any image‑processing algorithms.

Considering the choice of the moving‑average parameter $\alpha$,
the expected behavior is observed: smaller $\alpha$ values produce
smoother curves that respond only to significant changes and do so
gradually, whereas larger $\alpha$ values yield curves that more
closely follow the raw event rate and are more sensitive to minor
fluctuations. Examining multiple moving averages with different $\alpha$
values can therefore be useful for capturing various aspects of the
flow. For instance, $\alpha=0.001$ can be used to track overall trends
in the GaIn flow, $\alpha=0.01$ to detect rapid changes, and $\alpha=0.1$
to follow short‑term event variability.

For local monitoring, we selected $\alpha=0.01$ as a balanced intermediate
value. The observation window was divided into four vertical stripes,
as described in $\oint$\ref{subsec:Local-Monitoring}, ensuring an
even spatial distribution of events. We then tracked the exponential
moving averages ($\alpha=0.01$) within each stripe. As expected,
the moving averages for the individual stripes in Figure \ref{fig 5.2}
generally follow the same overall trends as the full‑window event
rates and their corresponding averages. However, they reveal an interesting
spatial dynamic: although the stripes begin with similar values, their
averages gradually diverge as the flow velocity decreases. Each stripe
contributes unevenly to the overall event rate and to the relative
distribution among stripes. In other words, flow instability is also
reflected in the growing inconsistency of the moving averages across
different regions.

\subsection{Scenario 2: Steady Flow Velocity\label{subsec:Scenario-2:-Constant}}

Figure \ref{fig 5.3} presents the global benchmarks for the scenario
in which the GaIn flow velocity was kept constant during an approximately
3.5‑minute recording. While the event rate initially appears stable,
closer inspection reveals a brief disturbance in the event flow around
the 65-second mark, followed by a gradual decline in the overall rate
after approximately 130 seconds. These irregularities suggest the
presence of potential anomalies in the system. 

\begin{figure*}[t]
\centering\includegraphics[scale=0.5]{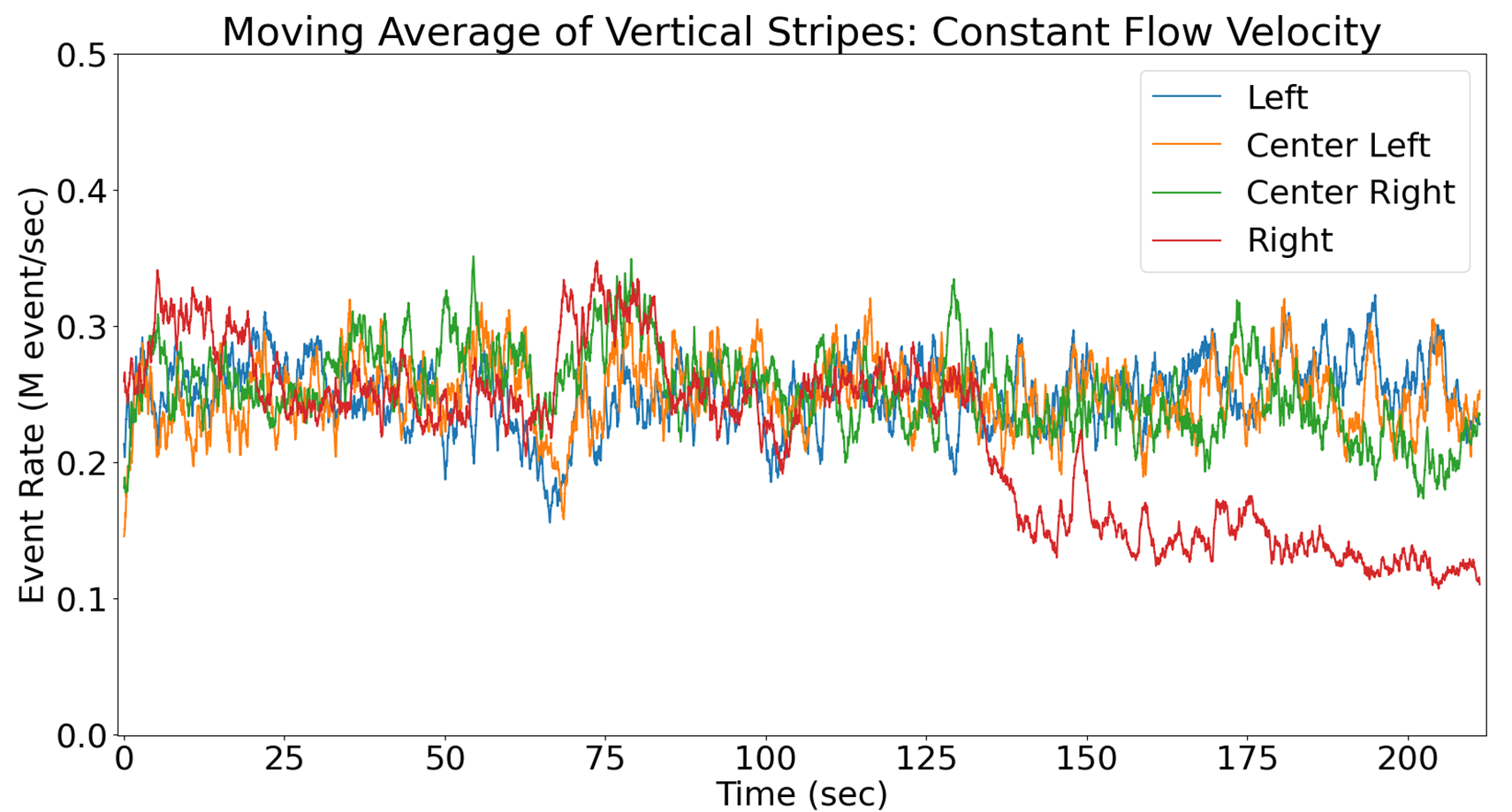}\caption{Exponential moving averages ($\alpha=0.01$) for the four vertical
stripes, showing a clear decrease in event rate in the rightmost stripe.}

\label{fig 5.4}
\end{figure*}

As shown in Figure \ref{fig 5.4}, a more detailed examination of
the local event rates indicates that the disturbance around 65 seconds
is visible across all stripes. In contrast, the decrease observed
after approximately 130 seconds can be attributed solely to a significant
drop in the event rate - nearly halving - in the rightmost stripe,
while the remaining stripes maintain relatively stable and comparable
rates throughout the recording. This gradual decline toward the end
highlights the capability of the method to detect subtle, slowly developing
phenomena in the flow before they evolve into more pronounced issues.

\section{Conclusion}

In this work, we introduced a novel framework for real-time monitoring
of gallium--indium flow based on the unique sensing paradigm of event
cameras. By representing the physical flow as an “event flow,” we
demonstrated that the temporal structure of the event rate provides
a direct and computationally efficient proxy for detecting flow dynamics
and anomalies, without the need for conventional image reconstruction
or optical flow estimation. We established a set of global and local
benchmarks, centered on the event-rate signal and its exponential
moving average, which enable both coarse and fine-grained analysis
of the flow behavior. A key contribution of this study is the development
of a practical heuristic for tuning event camera biases, including
the selection of an optimal time division constant and high-pass filter
setting, guided by the statistical properties of the signal. Experimental
results from the SARAF-II beam dump system show that the proposed
approach reliably tracks variations in flow velocity and reveals both
abrupt disturbances and slowly evolving spatial irregularities. These
findings highlight the potential of event-based sensing as a robust
and lightweight alternative for monitoring high-speed liquid metal
flows, and provide a foundation for future work on automated anomaly
detection and closed-loop control in demanding environments.

\pagebreak{}

\end{document}